\documentclass{article}
\usepackage{ijcai26}

\usepackage{times}
\usepackage{soul}
\usepackage{url}
\usepackage[hidelinks]{hyperref}
\usepackage[utf8]{inputenc}
\usepackage[small]{caption}
\usepackage{graphicx}
\usepackage{amsmath}
\usepackage{amsthm}
\usepackage{booktabs}
\usepackage{algorithm}
\usepackage{algorithmic}
\usepackage{float}
\usepackage[switch]{lineno}
\usepackage{xcolor}
\newcommand{\hspacehl}{\colorbox{blue!15}{\rule[0ex]{0pt}{0.8ex}\kern0.1em}}

\title{Say Anything but This: When Tokenizer Betrays Reasoning in LLMs}

\author{
Navid Ayoobi
\and
Marcus I Armstrong\and
Arjun Mukherjee\\
\affiliations
University of Houston\\
\emails
\{nayoobi, miarmstr\}@cougarnet.uh.edu,
amukher6@central.uh.edu
}

\begin{document}

\maketitle

\begin{abstract}
Large language models (LLMs) reason over discrete token ID sequences, yet modern subword tokenizers routinely produce non-unique encodings: multiple token ID sequences can detokenize to identical surface strings.
This representational mismatch creates an unmeasured fragility wherein reasoning processes can fail.
LLMs may treat two internal representations as distinct ``words'' even when they are semantically identical at the text level. 
In this work, we show that tokenization can betray LLM reasoning through one-to-many token ID mappings.
We introduce a tokenization-consistency probe that requires models to replace designated target words in context while leaving all other content unchanged.
The task is intentionally simple at the surface level, enabling us to attribute failures to tokenizer-detokenizer artifacts rather than to knowledge gaps or parameter limitations.
Through analysis of over 11000 replacement trials across state-of-the-art open-source LLMs, we find a non-trivial rate of outputs exhibit phantom edits: cases where models operate under the illusion of correct reasoning, a phenomenon arising from tokenizer-induced representational defects. 
We further analyze these cases and provide a taxonomy of eight systematic tokenizer artifacts, including whitespace-boundary shifts and intra-word resegmentation.
These findings indicate that part of apparent reasoning deficiency originates in the tokenizer layer, motivating tokenizer-level remedies before incurring the cost of training ever-larger models on ever-larger corpora.
\end{abstract}

\section{Introduction}
Recent breakthroughs in large language model (LLM) capabilities are predominantly attributed to scaling Transformer architectures with massive datasets and computational resources, yielding emergent abilities including reasoning, code generation, and few-shot generalization \cite{saunshi2025reasoninglatentthoughtspower,kaplan2020scalinglawsneurallanguage}. 
However, the prevailing ``scale explains everything'' narrative overlooks a foundational component that mediates all model input and output: the tokenizer \cite{gastaldi2025foundationstokenizationstatisticalcomputational}.
Tokenization constitutes the mandatory translation layer between the continuous, context-dependent nature of human language and the discrete symbolic representations required for computational processing. 
Early tokenizers employed word-level vocabularies, which suffered from out-of-vocabulary (OOV) limitations, where any word absent from the training corpus would be mapped to an unknown token, resulting in information loss \cite{Choo31122023}.
This limitation motivated a transition to open-vocabulary schemes. Accordingly, modern LLMs adopt subword tokenization including Byte Pair Encoding (BPE) \cite{sennrich-etal-2016-neural}, WordPiece \cite{wu2016googlesneuralmachinetranslation}, Unigram \cite{kudo-2018-subword}, and SentencePiece \cite{kudo-richardson-2018-sentencepiece}.
These algorithms decompose words into frequent substrings or character-level units, substantially solving OOV and rare word problem.
Yet subword tokenization also introduces new challenges. By defining how text is segmented, the tokenizer can impose structural artifacts on the model’s internal representation of language.
The availability of numerous subtoken units enables different subtoken combinations to concatenate into the same word, establishing representational non-uniqueness.
The core of our critique centers on the many-to-one nature of the tokenizer's mapping from token ID space to surface text. While the tokenization process is designed to be deterministic for a given input, the detokenization function is not injective.
Consequently, models must learn that different subtoken decompositions of the same word represent semantically identical content despite possessing different token-level encodings.
Simultaneously, research findings indicate that scaling model size and data volume alone does not resolve all reasoning deficiencies in LLMs.
State-of-the-art models exhibit remarkable performance across numerous benchmarks, yet they continue to exhibit systematic failures on certain reasoning tasks, particularly as problem complexity increases. 
This suggests that some limitations arise from factors beyond model scale, potentially from how information is represented or processed internally.

In this paper, we argue that tokenization is a hidden contributor to the reasoning failures. 
We introduce a simple yet diagnostic tokenization-consistency probe: models are instructed to perform straightforward word substitutions while preserving all other text unchanged.
Success requires identifying marked spans, applying designated edits, and preserving unmarked regions.
Under correct operation, an LLM should recognize that generating alternative token sequences which decode to the same word does not constitute a genuine edit and must be avoided.
However, our empirical results reveal an alarmingly high rate of tokenizer-induced ``phantom edits''.
In these cases, models output different token ID sequences for target words while the decoded surface text remains identical. 
The transformation appears successful from the model's internal token-based perspective (IDs differ), yet it is semantically null from the text-based evaluation perspective (strings match).
This demonstrates that models are systematically misled by tokenizer properties: they ``believe'' they have successfully executed substitutions when no actual content change has occurred.
Importantly, these failures do not reflect knowledge limitations or model scale, but rather expose a fundamental architectural constraint: \textbf{the inability to recognize that different token-level representations can encode identical linguistic content}.

Our analysis of over 11k replacement trials across ten state-of-the-art open-source LLMs reveals that a non-trivial fraction of reasoning failures stem not from knowledge limitations or model capacity constraints, but from tokenizer-induced representational non-uniqueness.
We further characterize the mechanisms underlying these failures through a taxonomy of eight systematic artifact types, including whitespace-boundary shifts, whitespace detachment and reattachment, newline substitutions, intra-word resegmentation, and proper-noun and morphological segmentation ambiguities. 
This taxonomy reveals how standard vocabulary redundancies and alternative segmentation pathways systematically mislead token-space reasoning, providing a roadmap for future tokenizer-aware architectural improvements.
Additionally, through a post-hoc token ID masking intervention, we demonstrate that when these spurious representational pathways are suppressed, models frequently reveal underlying reasoning capacities that were previously overshadowed. 
This suggests that the reasoning deficiency observed in state-of-the-art LLMs is partly a mirage, a byproduct of a mandatory translation layer that is both lossy and non-injective.
Ultimately, this research motivates a paradigm shift toward more robust representational schemes to ensure that future language models can reason as reliably at the surface level as they do within their latent embedding spaces.

\section{Related work}
\subsection{Tokenization in modern LLMs}
Subword tokenization techniques such as BPE \cite{sennrich-etal-2016-neural}, WordPiece \cite{wu2016googlesneuralmachinetranslation}, and Unigram \cite{kudo-2018-subword} were developed to address out-of-vocabulary (OOV) and rare word challenges.
These techniques form the tokenization backbone of modern state-of-the-art LLMs. 
Following the release of GPT-4 and the open-sourcing of OpenAI’s tiktoken library, the field broadly converged toward high-performance BPE variants.
Llama 3.2 and Qwen 3 employ Byte-Level BPE, an algorithm that constructs tokens from individual bytes to enable lossless encoding of multilingual text and code while mitigating the vocabulary explosion problem inherent in character-level BPE. 
This approach is further refined in the Mistral model family, which implements the ``Tekken'' tokenizer, a specialized BPE variant optimized for compression efficiency on source code and European languages.
In contrast to this BPE-dominated landscape, Gemma 3 adopts the Unigram language model implemented via SentencePiece \cite{kudo-richardson-2018-sentencepiece}, utilizing an extensive vocabulary exceeding 256k tokens to support massively multilingual capabilities and enable direct integration of visual embeddings into the text stream.
Despite their widespread success, fixed-vocabulary tokenizers have been criticized as a limiting inductive bias.
These limitations have motivated ``token-free'' approaches that compute directly on bytes or characters as the atomic units of digital text \cite{10.1162/tacl_a_00448,xue-etal-2022-byt5,tay2022charformerfastcharactertransformers}.

\subsection{Advances in LLM reasoning}
LLMs have achieved significant reasoning improvements \cite{Fang_Deng_Zhang_Shi_Chen_Pechenizkiy_Wang_2024} through structured prompting approaches, notably Chain-of-Thought (CoT) \cite{NEURIPS2022_9d560961,11123142,wang-etal-2023-towards} and Zero-Shot CoT \cite{NEURIPS2022_8bb0d291}, with subsequent work automating and refining these techniques through methods like Auto-CoT \cite{zelikman2024star,NEURIPS2022_9d560961}.
Beyond prompting, augmenting LLMs with external tools and APIs further enhances their reasoning capabilities \cite{gou2024toratoolintegratedreasoningagent,NEURIPS2023_871ed095,dong2025toolstarempoweringllmbrainedmultitool}.
Methods such as ReAct \cite{yao2023reactsynergizingreasoningacting} interleave logical reasoning steps with executable action commands, while frameworks like Toolformer \cite{NEURIPS2023_d842425e} enable models to invoke external resources including calculators, search engines, and specialized APIs, yielding improved performance on factual accuracy, numerical computation, and zero-shot reasoning tasks.
Parallel research on retrieval-augmented generation \cite{tran-etal-2025-rare} extends LLM capabilities by incorporating external knowledge bases that can be queried dynamically, effectively providing models with expandable memory beyond their parametric knowledge \cite{JMLR:v24:23-0037}.

\subsection{Tokenization as a reasoning bottleneck}
By fragmenting semantic continuity into discrete integer representations, tokenizers can constitute a bottleneck in LLM semantic understanding \cite{10.1162/tacl_a_00747}.
In quantitative reasoning, this appears as structural misalignment: greedy left-to-right subword tokenization can split integers and dates into arbitrary, non-semantic fragments, thereby obscuring the positional structure required for arithmetic and temporal inference \cite{singh2024tokenizationcountsimpacttokenization,bhatia2025datefragmentshiddenbottleneck,fatemi2024testtimebenchmarkevaluating}.
Manipulating digit-grouping tokenizations reveal substantial accuracy shifts and altered error distributions in state-of-the-art LLMs, indicating that model ``reasoning'' frequently adheres to tokenizer-imposed structure rather than learned abstract computational principles.
More broadly, small surface perturbations like typos, spacing changes, and rare Unicode characters can trigger different segmentations and produce behavioral changes \cite{chai-etal-2024-tokenization}.
Proposed mitigations include exposing models to multiple valid tokenizations during training \cite{provilkov-etal-2020-bpe}, and designing token-free architectures \cite{NEURIPS2024_e1f41845}.


\section{Method}
\subsection{Task setup overview}
\begin{figure*}[!t]
    \centering
    \includegraphics[width=\linewidth]{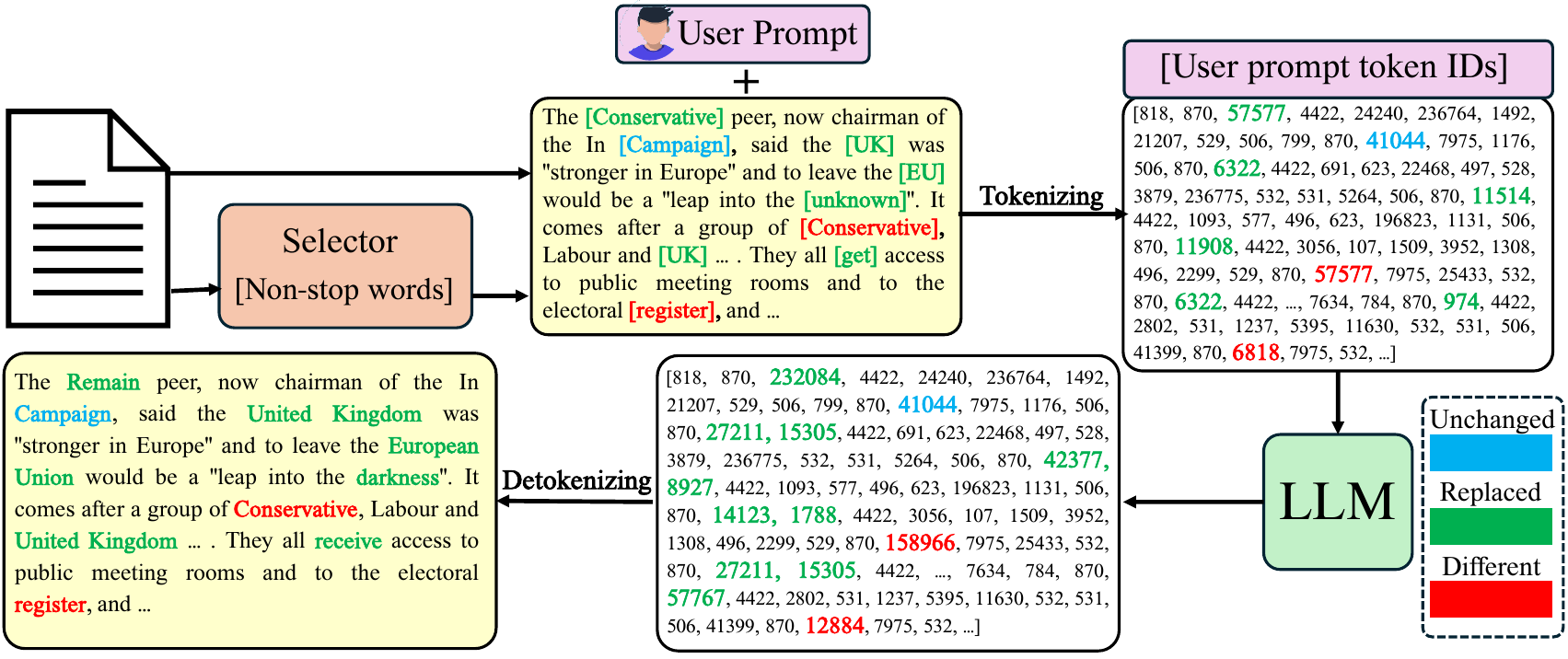}
    \caption{The overview of the proposed tokenization-consistency probe.}
    \label{fig:model}
\end{figure*}
LLMs execute instructions by operating over token ID sequences produced by tokenizers.
Tokenization poses a particularly strict challenge: the same surface word may be associated with different token IDs at input encoding versus output decoding, due to non-unique segmentations that detokenize to identical text.
This property establishes a natural framework for probing tokenization-consistency reasoning.
We define the task formally: given a text with marked spans and an instruction to replace each marked span with an alternative, the model must 1) identify the target spans, 2) apply the edits, and 3) preserve all non-target text.
Our probe evaluates whether models recognize that different token ID combinations can decode to identical strings and therefore represent identical semantics despite divergent internal representations. 
This setting isolates failures arising from the interaction between the tokenizer's discrete alphabet and the model's learned rules, thereby evaluating reasoning under the tokenizer’s lossy and non-unique encoding.
In contrast to standard benchmarks where reasoning failures may reflect insufficient model parameters, knowledge gaps, or sampling variance, our task specifies an unambiguous objective where failures arise because the model's internal processing depends on token IDs that lack one-to-one correspondence with surface text.

Figure \ref{fig:model} depicts the full experimental workflow.
The selector randomly samples $p\%$ of non-stop words from each document and marks them by placing them inside brackets. 
We then prepend an explicit replacement instruction to the document and pass the combined text through the tokenizer.
The tokenizer maps the input to a token-ID sequence that serves as the LLM's input.
Finally, the model's output token IDs are detokenized to produce the output string.
We evaluate replacement performance by comparing input and output texts together with their aligned token-ID sequences.
Using evidence from both the surface form and the token IDs, we categorize model behavior into three classes:
\textbf{1) \textit{Unchanged}}, where the model does not replace the target and the output token ID matches the input token ID; 
\textbf{2) \textit{Replaced}}, where the model performs a successful substitution and the output token ID differs from the input token ID; and
\textbf{3) \textit{Different}}, where the model outputs different token ID(s) that nevertheless decode to the same word, yielding the same surface string as the original input word, despite token-id variation.
In the example shown in Figure \ref{fig:model}, the model successfully replaces several targets (e.g., get $\rightarrow$ receive), indicated in green.
As expected, successful substitutions are accompanied by changes in the corresponding token IDs.
Conversely, the model failed to replace the word ``Campaign'', which retains the same token ID in the output sequence, indicated by blue highlighting.
Most critically, the red color highlights the cases where the model outputs a different token-ID representation than the input, but detokenization produces the same surface word in the final text.
This phenomenon arises because tokenization is not injective, distinct token-ID sequences can map to an identical surface form.
We provide a more detailed explanation of how the tokenizers enable such token-level changes without surface-level edits in Section \ref{sec:errors}.

\subsection{Experimental setup}
\textbf{Dataset:} we sample news articles from the XSUM dataset \cite{narayan2018dontdetailsjustsummary} and retain texts between 100 and 600 words.
For each selected article, we randomly sample $5\%$ of non-stop words as replacement targets and designate these spans by enclosing them in brackets, producing a bracket-annotated version of the original text. 
LLMs receive the bracket-annotated document along with explicit instructions to replace only bracketed words while maintaining all other text in its original form.
\\
\textbf{LLM models:} we evaluate ten state-of-the-art open-source LLMs spanning four major model families and accessed through Hugging Face: Gemma (Gemma3-270M, Gemma3-1B, Gemma3-4B, Gemma3-12B), Llama (Llama3.2-1B, Llama3.2-3B, Llama3.1-8B), Mistral (Ministral-8B, MistralSmall-24B), and Qwen (Qwen3-4B, Qwen3-30B).
To ensure comparability across models, we employ a standardized sampling configuration for all generation tasks, and set Top-p, Top-k, and Temperature to 0.9, 50, and 1.0, respectively.

\section{Result and discussion}
\subsection{Model size is not always the bottleneck for reasoning}
\begin{figure*}[!ht]
    \centering
    \includegraphics[width=0.8\linewidth]{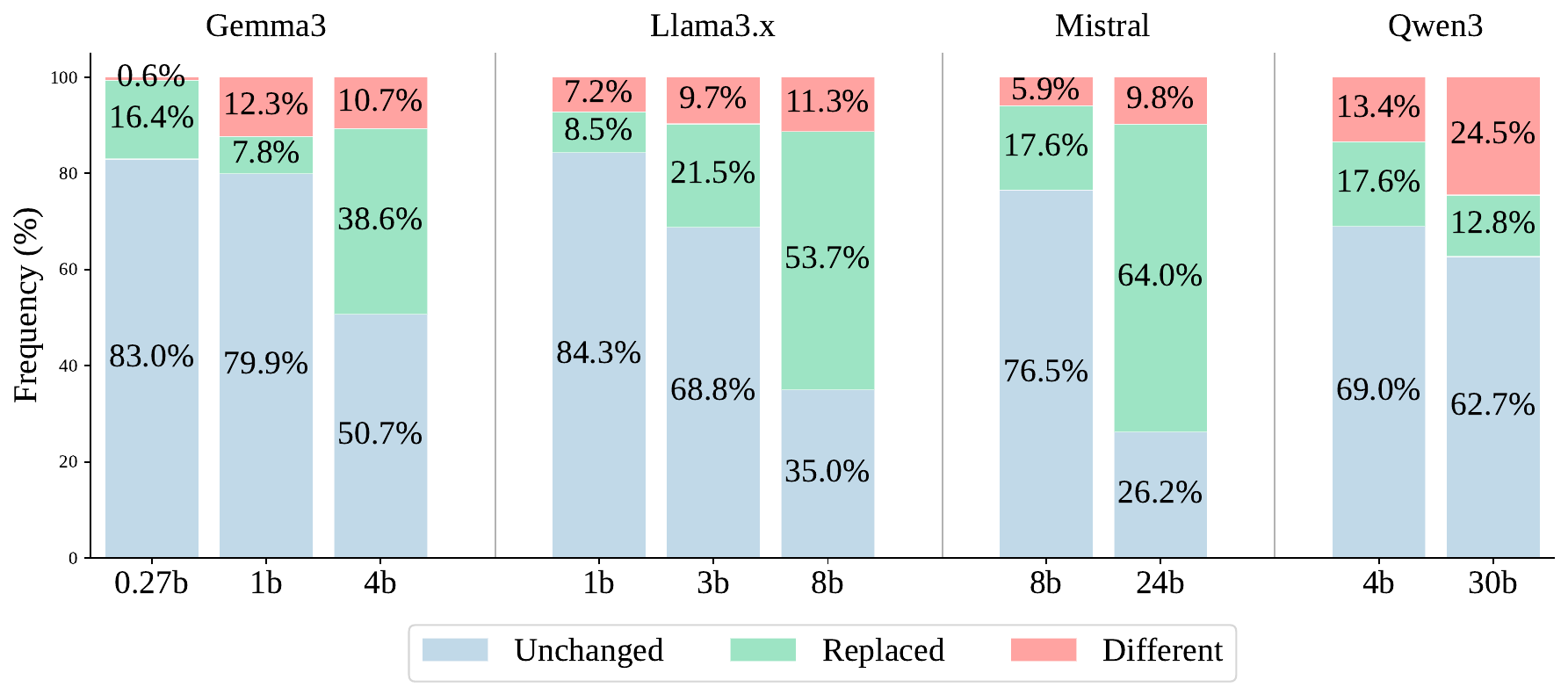}
    \caption{Distribution of outcomes across LLM families and parameter scales for the word-replacement task. The ``Different'' category (red) highlights tokenizer-induced phantom edits, which persist across all model sizes and families.}
    \label{fig:parameters}
\end{figure*}

While parameter scaling is often assumed to improve reasoning capacity, our experiments show that it is not always sufficient to ensure semantic understanding and reasoning in targeted word-replacement tasks.
Figure \ref{fig:parameters} aggregates outcomes from over $11$k replacement trials, categorizing model performance into our three distinct classes (\textit{Unchanged}, \textit{Replaced}, and \textit{Different}).

If parameter insufficiency were the primary failure mode, we would anticipate monotonic performance improvements with model size within a given architecture family.
However, our observations contradict this expectation. 
Within the Qwen3 and Gemma3 families, some larger variants achieve replacement success that is comparable to or even lower than that of smaller counterparts.
Moreover, even in cases where larger models behave better overall (higher \textit{Replaced} and lower \textit{Unchanged} rates), a non-trivial portion of outputs still fall into the \textit{Different} category.
These outcomes are consistent with a tokenizer-induced representational defect.
For some word forms, the mapping from token-ID sequences to rendered text is effectively many-to-one, such that different subtoken segmentations decode to the same surface string.
Consequently, models may execute legitimate substitutions at the token level (replacing one valid tokenization with an alternative) while producing no perceptible modification at the character level, creating the appearance of replacement failure.
The model therefore ``believes'' the replacement occurred due to token-ID changes, even though the decoded output remains identical.
Critically, because this discrepancy originates in the tokenizer-detokenizer architecture rather than in the model's parametric representation, increasing model capacity offers no systematic solution to this fundamental misalignment.
\subsection{Tokenizer-induced phantom edits} \label{sec:errors}
We examine token-level input-output differences within the \textit{Different} class to identify systematic tokenizer artifacts that can mislead the model into believing it reasons correctly for word replacement.
These errors arise from the non-unique representational property inherent to subword tokenization schemes, wherein a single surface string may admit multiple valid segmentations.
This structural ambiguity establishes a many-to-one mapping from token ID sequences to character strings.
Therefore, models may modify token IDs through boundary shifts, whitespace insertion/deletion, or alternative segmentations yet the final decoded text remains identical.
We refer to such cases as phantom edits, syntactically valid token-space modifications that collapse to an unchanged surface form, obscuring the semantic equivalence across segmentations and hindering the model's ability to reason reliably about word identity. Below, we present our taxonomy of these errors.

\subsubsection{Error 1: whitespace-boundary shift}
A prominent error mode stems from tokenizer's vocabulary that duplicate a word with (shown in this paper as \hspacehl) and without a leading whitespace (e.g., ``February'' vs. ``\hspacehl February''), which are assigned distinct token IDs and thus distinct representations.
In this mode, models substitute a space-prefixed token with its non-prefixed counterpart, producing a token ID change that decodes to the same surface string after whitespace normalization during detokenization.

\subsubsection{Error 2: whitespace detachment/reattachment}
We identify two related error modes driven by whitespace-sensitive vocabulary design.
First, a token with leading space like ``\hspacehl Saturday'' is converted into an explicit space token together with the same word token, e.g., [``\hspacehl'', ``\hspacehl Saturday''], which we term \textbf{whitespace reattachment}.
Second, a token such as [``\hspacehl Guy''] is split into an explicit space token and the word token without a leading space, [``\hspacehl'', ``Guy''], which we call \textbf{whitespace detachment}.
Both errors originate from the tokenizer vocabulary's redundant encoding of words, wherein both space-prefixed and non-prefixed versions of the same word exist as distinct entries, enabling multiple valid segmentations that decode to identical surface strings.

\subsubsection{Error 3: newline/whitespace substitution}
We also find cases where a space-prefixed token is transformed into a newline token plus the space-free word token, as in [``\hspacehl However'']$\rightarrow$[``\texttt{\string\n}'', ``However'']. 
From the model's representational perspective, this constitutes a valid replacement operation: the output token IDs differ from the input, signaling successful task completion. 

\subsubsection{Error 4: intra-word resegmentation}
During tokenization, the tokenizer often prefers a single vocabulary entry that fully matches the word (i.e., an unsplit token).
However, because the vocabulary contains multiple subword units that can be compositionally combined to form the same lexical item, models may substitute an atomic token with a multi-token sequence comprising its constituent subpieces, unaware that this decomposition yields an identical surface string upon decoding. 
For instance, when replacing ``\hspacehl unbelievable'', the model may generate the three-token sequence [``un'', ``bel'', ``ievable''].
This transformation converts a single token ID [66917] into a sequence of distinct IDs [602, 8145, 179086], producing entirely different token identifiers and consequently divergent vector representations, despite maintaining surface-level equivalence.

\subsubsection{Error 5: proper-noun segmentation ambiguity}
Proper nouns and rare words are typically tokenized through subword decomposition, as atomic representations often do not exist in the vocabulary, for example, [``\hspacehl Jub', ``ilee''].
During inference, models may substitute the original segmentation with alternative valid decompositions that redistribute characters across token boundaries, such as [``J'', ``ub'', ``ilee''].
Our analysis also reveals a systematic pattern wherein models isolate initial capital letters as separate tokens before applying subword segmentation to the remainder, for instance, transforming [``\hspacehl Dorm'', ``er''] into [``D'', ``orm'', ``er'']. 
Notably, while these alternative segmentations may share some common token IDs (e.g., ``er''), the overall token ID sequences differ from the input, leading the model to interpret this resegmentation as a successful replacement despite producing identical decoded output.

\subsubsection{Error 6: morphological boundary surfacing}
Tokenizers exploit morphological regularities by representing common affixes as distinct tokens, enabling compositional word construction, for instance, segmenting [``\hspacehl repaid''] as [``re'', ``paid''].
However, because such morphemic pieces appear across many lexical items, they are not uniquely tied to the intended word-level meaning (e.g., ``paid'' in this context).
Consequently, when models perform morphological resegmentation during replacement, they may erroneously interpret the combination of generic morphemic tokens as constituting semantically distinct content from the original atomic word, despite their decode-time equivalence.

\subsubsection{Error 7: acronyms split}
Acronyms behave differently from ordinary words because they can originate from multiple underlying expansions and often admit several valid tokenizations that detokenize to the same surface string.
Since many tokenizers include subtokens for single capital letters as well as multi-letter capital sequences, the model may assemble an acronym using different capitalized units and treat this as a meaningful change in its internal reasoning, even though the decoded acronym is unchanged (e.g., ``\hspacehl HIV''$\rightarrow$[``H'', ``IV'']).
Note that we do not expect acronyms, often uniquely grounded in a document, to be replaced with different acronyms.
Instead, a reasonable replacement would be to expand them into their full phrases (like the example presented in Figure \ref{fig:model}).
However, the predominant pattern involves resegmentation among all-capital token variants, where distinct token ID sequences mislead the model into inferring it has generated new lexical content despite surface-level equivalence.

\subsubsection{Error 8: plural/possessive tail tokens}
The single-letter token `s', like other isolated alphabetic characters, lacks inherent semantic content and functions primarily as a compositional subtoken within larger units, for example, ``\hspacehl smooth'' may be segmented as [``s'', ``mooth''].
As a consequence, plural forms and possessives may be resegmented into a base token plus the isolated ``s'', producing token-ID sequences that differ substantially from the original unsplit tokenization (e.g., ``\hspacehl rights''$\rightarrow$[``right'', ``s''], and [``\hspacehl Clement'', ``s'']$\rightarrow$[``C'', ``lements'']).
The resulting token-space divergence can mislead the model into treating the output as a meaningful modification, thereby reinforcing the tokenizer fault and obscuring erroneous reasoning about word identity.

This phenomenon underscores a critical tokenizer design limitation: the morphological marker `s' (denoting possession or plurality) and the alphabetic character `s' (functioning as a compositional subtoken in larger units) are conflated into a single vocabulary entry despite serving fundamentally different linguistic roles. 
This representational conflation enables the problematic resegmentations described above, as the tokenizer cannot distinguish between morphological and compositional uses of the same character sequence, permitting models to substitute one role for another while remaining unaware of the semantic distinction.

\subsection{Subtokens' realignment patterns}
\begin{figure}[ht]
    \centering
    \includegraphics[width=0.95\linewidth]{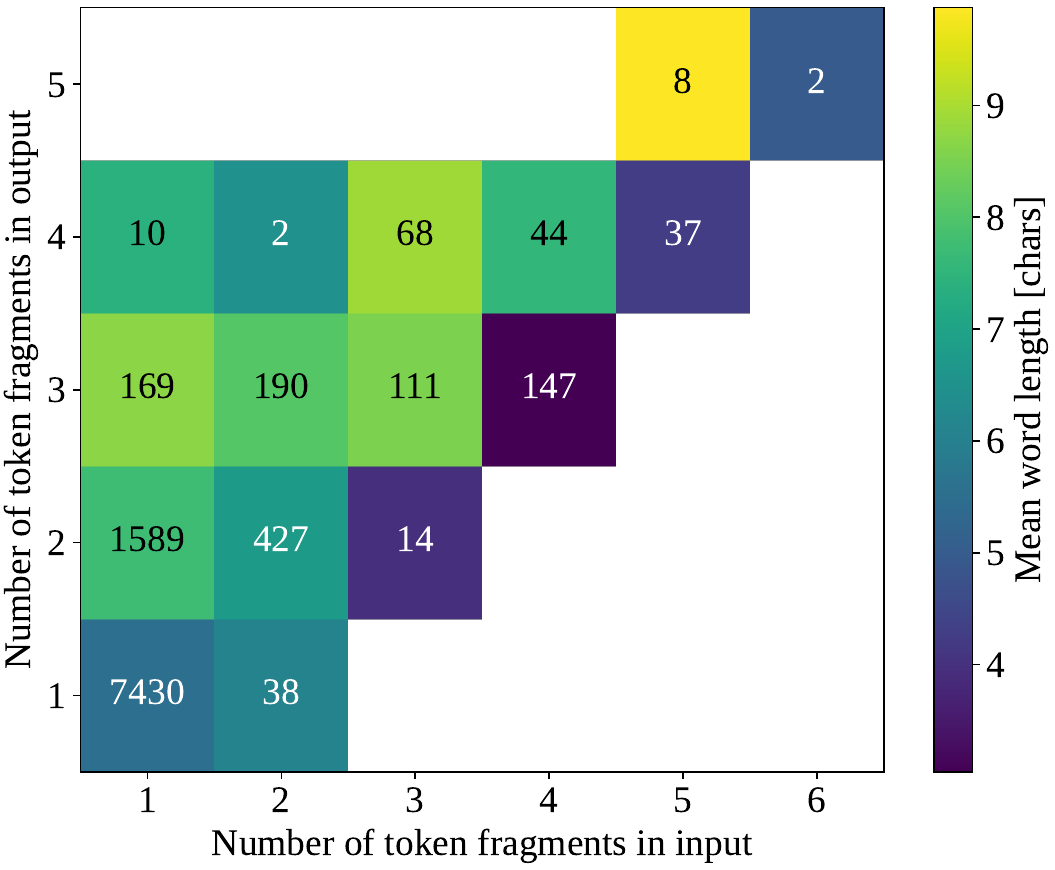}
    \caption{Heatmap of subtoken fragmentation transitions for ``Different'' class instances using the Gemma3-4B model. The x and y axes denote the number of token fragments in the input and output, respectively. Cell values indicate the frequency of each transition, while the color scale represents the mean character length of the words.}
    \label{fig:heatmap}
\end{figure}
\begin{figure*}[ht]
    \centering
    \includegraphics[width=0.8\linewidth]{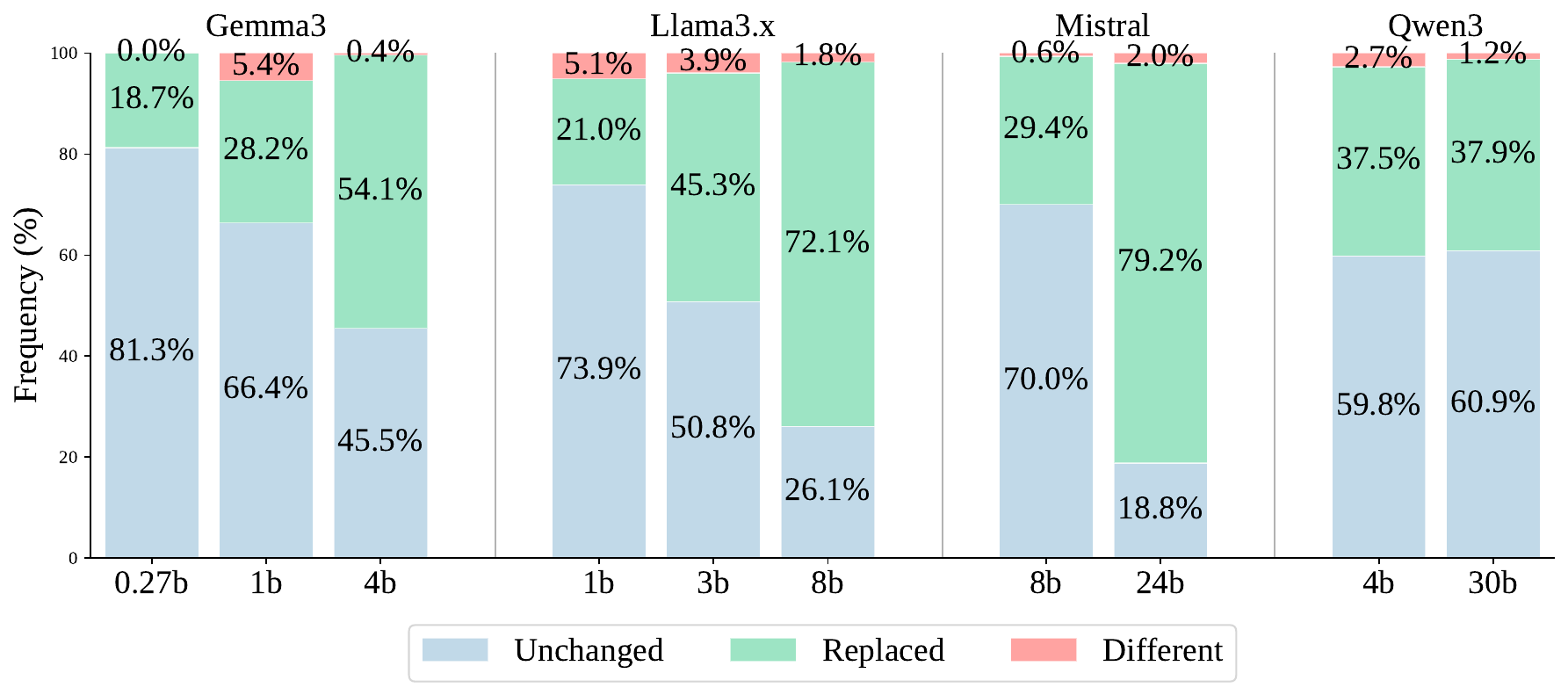}
    \caption{Distribution of outcomes across LLM families after applying the token-ID masking intervention.}
    \label{fig:remove}
\end{figure*}
In this experiment, we sample non-stop words with character lengths ranging from 3 to 15 characters, mark them within news articles, and present these annotated documents to Gemma3-4B model. 
Our analysis centers on the ``Different'' class exclusively.
For each instance in this class, we quantify two values: the number of subtokens produced by the tokenizer for the input target word, and the number of subtokens used for the corresponding surface word in the model output.
we aggregate these pairs into the 2D grid of Figure \ref{fig:heatmap}.
Each cell represents a specific fragmentation transition (e.g., 1${\rightarrow}$2 subtokens).
We label each cell with the frequency of ``Different'' instances mapped to that transition and use a color scale to represent the average character length of the words in that cell.

\textbf{Finding 1.} Cells above the diagonal line indicate that subtoken splitting (increasing subtoken count) is common while merging (decreasing subtoken count) is rare.
Aggregating across the grid, we observe that approximately $78\%$ of cases preserve the same subtoken count ($y{=}x$), about $19.7\%$ increase the count ($y{>}x$, representing splits), and only $2.3\%$ decrease it ($y{<}x$, representing merges).
This pronounced asymmetry suggests that models more frequently edit by expanding representations than by compressing them.
This pattern raises two critical questions.
First, if a more concise tokenization exists, why does the tokenizer sometimes assign a more fragmented input encoding (below-diagonal cases)?
Second, why does the model more frequently expand representations at inference than compress them, effectively ``over-complicating'' the representations (above-diagonal cases)?
These patterns suggest that LLMs rarely succeed in ``repairing'' suboptimal segmentations, introduced upstream by the tokenizer.

\textbf{Finding 2.} Most input words possess atomic token representations, meaning they have unique IDs in the tokenizer vocabulary and theoretically require no subtoken decomposition. 
However, this theoretical property does not hold in practice: words are frequently fragmented into valid subtokens during inference despite having atomic representations available.
Conversely, the existence of duplicate single-token variants causes systematic model failures. Notably, the (1$\rightarrow$1) cell is the single most common transition, accounting for approximately $72.2\%$ of all ``Different'' instances.
This reveals that the majority of phantom edits preserve subtoken count while altering only token ID identity.
This pattern reveals that the substantial existence of whitespace-variant token pairs (Error Type 1), tokens differing solely in leading whitespace, constitutes a primary failure mechanism. 
Models fail to recognize that these variants encode identical semantics, implying that their learned vector representations are not sufficiently close in embedding space to support equivalence recognition.

\textbf{Finding 3.} Mean character length generally increases with segmentation depth, though the relationship is not strictly monotonic.
The brightest region appears near the (5$\rightarrow$5) cell, consistent with the intuition that longer words often require more subtokens.
However, certain high-token-count cells exhibit darker coloring (indicating shorter mean character length), supporting the hypothesis that token-count inflation can arise from tokenizer artifacts rather than lexical complexity.
Examples include whitespace tokens, newline characters, and morpheme decompositions (with extra white spaces), non-lexical structural elements that increase token count without proportionally increasing character length.

\subsection{Superficial remedy via token-ID masking}
In practice, rectifying tokenizer deficiencies would require modifying tokenization rules and vocabulary, followed by re-training model so it can internalize the revised token-ID mappings. 
Token embeddings, output projections, and internal activation statistics are co-adapted to the original segmentation and ID assignments; thus, simply swapping token IDs or altering merge rules without re-optimization typically disrupts the learned mapping between subword units and semantics, leading to broad performance degradation.
Complete tokenizer redesign and LLM re-training remain largely inaccessible in academic settings due to computational constraints.
Therefore, we adopt a lightweight post-hoc intervention.
We mask the specific output token IDs that empirically trigger phantom edits in ``Different'' instances. 
This procedure does not modify the tokenizer, the embedding space, or the model’s learned representations.
It only constrains the decoding space so the model cannot assign probability mass to the offending tokens observed in ``Different'' samples, thereby forcing alternative generations.
We interpret this approach as a superficial remedy, a diagnostic and mitigation layer that suppresses a known failure mode without resolving the underlying non-uniqueness and tokenizer-model co-adaptation.
This intervention provides dual value: it 1) provides evidence that a meaningful fraction of errors is mediated by tokenization artifacts rather than model capacity, and 2) provides a low-cost workaround when retraining with a revised vocabulary is infeasible.

Figure \ref{fig:remove} presents LLM performance following the blocking of problematic token IDs.
Compared to the baseline shown in Figure \ref{fig:parameters}, the updated distribution reveals dramatic suppression of the ``Different'' failure mode, with percentages declining to approximately $0{-}5\%$ across all models.
However, the persistence of small ``Different'' percentages indicates that alternative token ID sequences still deceive models, which fail to recognize identical underlying semantics. 
This confirms that removing undesired token sequences cannot guarantee complete resolution of the issue, as the fundamental many-to-one mapping property of tokenization remains.
Interestingly, blocking problematic token IDs also produces reductions in ``Unchanged'' rates for most LLMs. 
This effect is consistent with decoding dynamics: suppressing a subset of high-probability ``phantom-edit'' tokens forces the model to renormalize and redistribute probability mass across remaining options, which can elevate truly correct replacements and increase the chance of producing a valid substitution.

Overall, the results indicate that once tokenizer-driven artifact tokens are blocked, models more reliably produce semantically distinct substitutions.
This demonstrates that models inherently possessed the reasoning capacity to generate appropriate replacements, but this capacity was systematically suppressed by the presence of high-probability tokenizer artifacts. 
This suggests that tokenizer-induced phantom edits constitute a ``path of least resistance'' in the probability landscape.
When this path is blocked, models show decreased input copying (``Unchanged'') and are instead forced to engage genuine semantic reasoning processes, successfully identifying alternative words that were previously overshadowed by the substantial probability mass allocated to phantom tokens.

\section{Conclusion and future work}
In this paper, we demonstrate that a non-trivial fraction of reasoning failures in LLMs originates upstream of the model, stemming from the tokenizer’s non-injective mapping between token-ID sequences and surface strings.
Using a simple tokenization-consistency probe, we expose ``phantom edits'', cases where models alter token IDs yet produce detokenized outputs identical to the input, thereby internally ``believing'' an edit has occurred despite null surface-level change.
Across evaluated open-source LLM families, these failures persist even as model parameters increase, indicating that scaling alone cannot systematically resolve reasoning deficiencies rooted in tokenization-induced representational mismatches.
We provide a systematic taxonomy of the mechanisms underlying these failures, including whitespace-boundary manipulations and intra-word resegmentation patterns, elucidating how standard vocabulary redundancies and alternative segmentation strategies systematically compromise LLM reasoning.

A plausible next step is to test whether tokenization non-uniqueness affects reasoning behaviors beyond our replacement probe.
Concretely, one could construct controlled ``equivalence interventions'' that replace an input's token ID sequence with an alternative sequence detokenizing to the identical surface string, then evaluate whether downstream behaviors (like chain-of-thought stability, multi-step arithmetic accuracy) exhibit sensitivity to these representationally distinct but semantically identical inputs.
Such experiments would quantify the extent to which reasoning pipelines are sensitive to representational choice rather than semantics, helping to separate genuine cognitive limitations from artifacts introduced by the tokenizer’s many-to-one mapping.
A second research direction involves tokenizer-aware training objectives that enforce representational consistency across equivalent encodings.
This can be achieved by modifying training procedures to use averaged (or pooled) representations across multiple token ID sequences corresponding to the same surface token or string, encouraging models to internalize their equivalence and reduce phantom-edit pathways. 
Additionally, the tokenizer can be redesigned to shrink equivalence classes, for example, introducing a dedicated $\mathtt{<start{\_}of{\_}sentence>}$ token for no-leading-space variants can eliminate duplicate forms of the same word (with vs. without leading space) and shrink the equivalence classes that enable spurious token-level edits (``token''$=$ [$\mathtt{<start{\_}of{\_}sentence>}$,``\hspacehl token'']).

\clearpage
\bibliographystyle{named}
\bibliography{ijcai26}

\end{document}